\documentclass{article}
\usepackage[preprint]{neurips_2024}

\author{
Nanbeige LLM Lab, Boss Zhipin
}

\usepackage{neurips_2024}

\usepackage{booktabs}
\usepackage{makecell}  

\usepackage[utf8]{inputenc} 
\usepackage[T1]{fontenc}    
\usepackage{hyperref}       
\usepackage{url}            
\usepackage{booktabs}       
\usepackage{amsfonts}       
\usepackage{nicefrac}       
\usepackage{microtype}      
\usepackage{xcolor}         
\usepackage{minted}
\usepackage{tcolorbox}
\tcbuselibrary{listings, skins}
\tcbuselibrary{breakable}
\usepackage{algorithm}
\usepackage{algpseudocode}
\usepackage{amsmath}
\usepackage{url}
\usepackage{graphicx}
\usepackage{amsmath}
\usepackage{amsthm}
\usepackage{booktabs}
\usepackage{color}
\usepackage{xcolor}
\usepackage{xspace}
\usepackage[misc]{ifsym}

\usepackage[ruled, vlined, nofillcomment, linesnumbered, algo2e]{algorithm2e}

\usepackage{booktabs}
\usepackage{multirow}
\usepackage{balance}
\usepackage{enumitem}
\usepackage{stfloats}
\usepackage{diagbox}
\usepackage{fancyhdr}
\usepackage{stfloats}
\definecolor{result_color}{RGB}{250,250,210}
\usepackage{bm}

\newcommand{\ignore}[1]{}
\newcommand{\paratitle}[1]
{\vspace{1.5ex}\noindent\textbf{#1}}

\title{Nanbeige4-3B Technical Report: Exploring the Frontier of Small Language Models}

\begin{document}
\noindent\includegraphics[height=0.8cm]{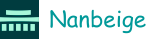}

\maketitle

\begin{abstract}
We present Nanbeige4-3B, a family of small-scale but high-performing language models. Pretrained on 23T high-quality tokens and finetuned on over 30 million diverse instructions, we extend the boundary of the scaling law for small language models.
In pre-training, we design a Fine-Grained Warmup-Stable-Decay (FG-WSD) training scheduler, which progressively refines data mixtures across stages to boost model performance. In post-training, to improve the quality of the SFT data, we design a joint mechanism that integrates deliberative generation refinement and chain-of-thought reconstruction, yielding substantial gains on complex tasks. 
Following SFT, we employ our flagship reasoning model to distill Nanbeige4-3B through our proposed Dual Preference Distillation (DPD) method, which leads to further performance gains. Finally, a multi-stage reinforcement learning phase was applied, leveraging verifiable rewards and preference modeling to strengthen abilities on both reasoning and human alignment.
Extensive evaluations show that Nanbeige4-3B not only significantly outperforms models of comparable parameter scale but also rivals much larger models across a wide range of benchmarks.
The model checkpoints are available at \url{https://huggingface.co/Nanbeige}.

\end{abstract}

\begin{figure}[H]
	\centering
	\includegraphics[width=1.0\textwidth]{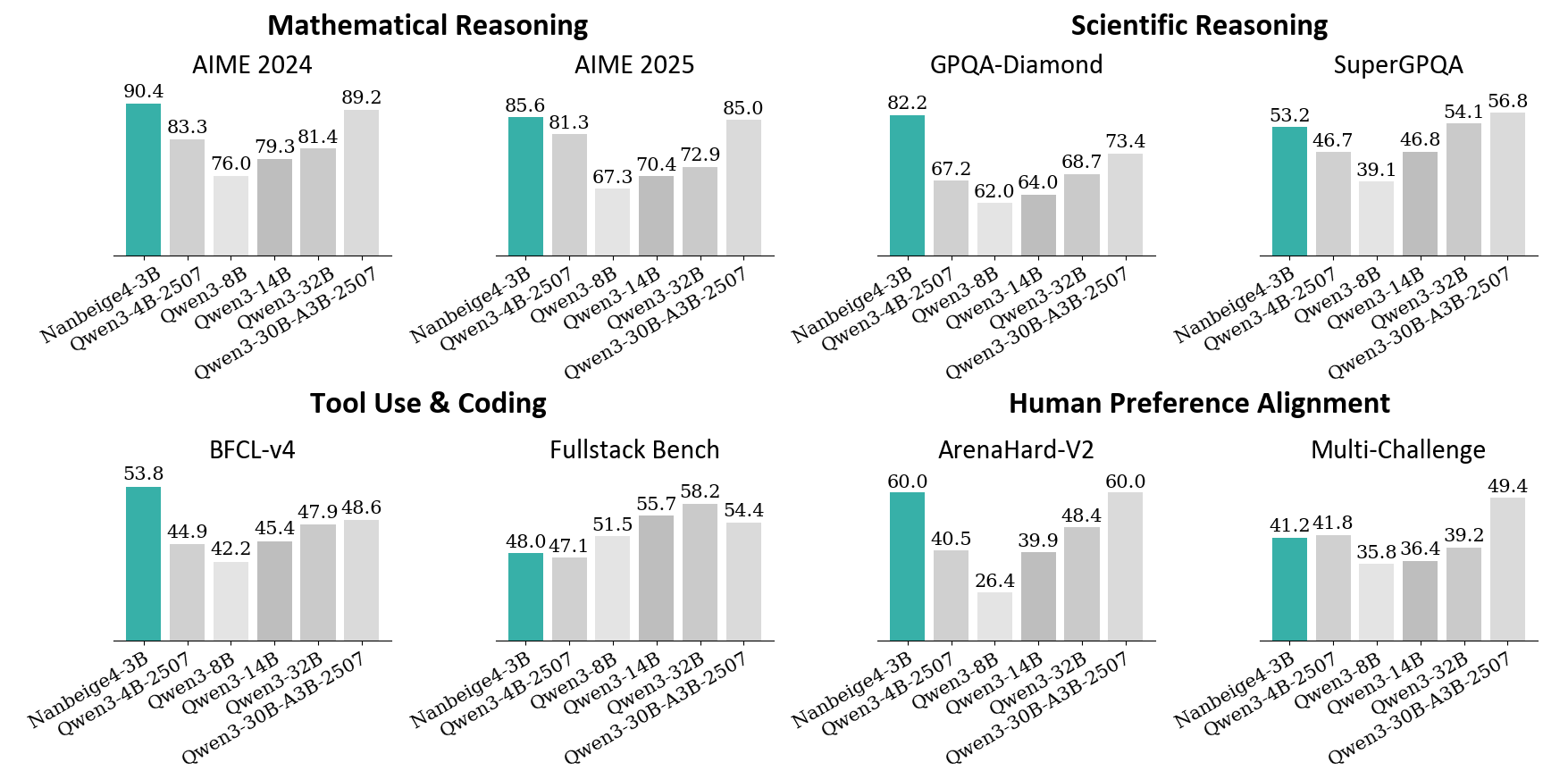}
	\caption{\textcolor{black}{Performance Comparison between Nanbeige4-3B-Thinking and Qwen series models.
 }}
	\label{fig:thinking_performance}
\end{figure}
\section{Introduction}\label{sec:intro}

In recent years, we have witnessed the emergence of capable large-scale language models ranging from hundreds of billions to trillions of parameters~\cite{yang2025qwen3, DBLP:journals/corr/abs-2508-06471, deepseekai2025deepseekr1incentivizingreasoningcapability, lingteam2025activationboostedscalinggeneral}. 
They have demonstrated remarkable reasoning abilities and have become a pivotal driving force in the evolution of artificial intelligence. Despite their impressive performance, these models come with substantial inference costs in deployment and high training expenses for research, whether for full-scale replication or fine-tuning. This scenario underscores the importance of exploring the potential of Small Language Models (SLMs) as a resource-efficient alternative.

In this work, we present the Nanbeige4-3B model family. Despite their compact size, they exhibit remarkably strong and well-balanced capabilities in mathematics, scientific reasoning, human preference alignment, creative writing, and tool use—substantially outperforming many larger models. These results highlight the effectiveness of our training methodology and clearly demonstrate that \textbf{small, well-engineered models can achieve performance that surpasses far larger models.}

To provide a clear sense of the model’s capability, we compare Nanbeige4-3B against the Qwen3 series across multiple parameter scales. For the base models, we conduct several post-training runs using identical SFT datasets, where Nanbeige4-3B-Base significantly outperforms Qwen3-8B-Base. For the reasoning model, Nanbeige4-3B-Thinking demonstrates overall performance that surpasses Qwen3-14B on average. Notably, it even outperforms the substantially larger Qwen3-32B and Qwen3-30B-A3B on mathematical and scientific reasoning, as well as on specific dimensions concerning tool use and human preference alignment. Moreover, in terms of creative writing ability, Nanbeige4-3B-Thinking approaches the performance of several state-of-the-art large models, as elaborated in the WritingBench Leaderboard November 2025.

\begin{table}[!h]
\caption{Performance comparison between Nanbeige4-3B-Thinking and other large parameter-scale models on WritingBench~\cite{wu2025writingbenchcomprehensivebenchmarkgenerative}. Scores are taken from the official \href{https://huggingface.co/spaces/WritingBench/WritingBench}{leaderboard}.}
\label{tab:writingbench}
\centering
 \small
\setlength{\tabcolsep}{3pt}
\resizebox{1 \textwidth}{!}{
\begin{tabular}{l c c c c c c c}
\toprule
Model & Overall &
\makecell[c]{Academic \&\\Engineering} &
\makecell[c]{Finance \&\\Business} &
\makecell[c]{Politics \&\\Law} &
\makecell[c]{Literature \&\\Arts} &
Education &
\makecell[c]{Advertising \&\\Marketing} \\
\midrule
GPT-5               & 83.87 & 84.46 & 83.78 & 82.21 & 85.26 & 85.64 & 82.34 \\
Qwen-235B-A22B-Thinking       & 82.34 & 83.43 & 83.44 & 79.67 & 82.68 & 84.52 & 80.94 \\
Doubao-Seed-1.6-Thinking  & 79.29 & 80.79 & 79.74 & 78.05 & 78.37 & 81.48 & 77.95 \\
Gemini-2.5-Pro                   & 79.26 & 79.59 & 78.24 & 78.69 & 80.92 & 81.30 & 77.27 \\
\underline{\textbf{Nanbeige4-3B-Thinking-2511}}    & 79.03 & 80.77 & 82.37 & 77.41 & 76.05 & 80.81 & 76.55 \\
 Deepseek-R1-0528                 & 78.92 & 78.79 & 78.38 & 77.27 & 80.91 & 80.41 & 78.44 \\
Grok-4               & 74.65 & 75.44 & 73.99 & 73.08 & 74.66 & 77.48 & 74.73 \\
O4-mini                 & 72.90 & 76.04 & 73.19 & 71.26 & 69.87 & 75.91 & 72.66 \\
O3-mini                 & 68.02 & 69.52 & 68.61 & 66.92 & 66.54 & 71.48 & 65.95 \\

\bottomrule
\end{tabular}
 }
\label{tab:model-domain-scores}
\end{table}

In this report, we present a comprehensive overview of the model training pipeline—from the pre-training to the post-training process, and highlight the techniques that most substantially contribute to our model’s performance improvements.
Specifically, Nanbeige4-3B is featured with the following technologies to improve model capabilities:

\paratitle{Pre-training with Hybrid Data Filtering and Fine-Grained WSD.}

\begin{itemize}[leftmargin=10pt]
    \item \textbf{Hybrid Data Filtering.} To enable more precise filtering of high-quality data, we develop a hybrid strategy combining tagging-based scoring~\cite{su2025nemotroncctransformingcommoncrawl, zhao2024decoratelmdataengineeringcorpus} with retrieval-based recalling~\cite{DBLP:journals/corr/abs-2401-14624}, obtaining the final comprehensive training corpus consists of 23 trillion tokens high-quality data. 

    \item \textbf{Fine-Grained WSD.} We introduce a Fine-Grained Warmup-Stable-Decay (FG-WSD) scheduler to maximize the utility of high-quality data. This approach employs a fine-grained and quality-progressive data curriculum, in which the stable stage is partitioned into multiple phases with progressively refined data mixtures. Evaluations demonstrate that our method yields substantial improvements over the vanilla WSD scheduler~\cite{hu2024minicpmunveilingpotentialsmall, liu2024deepseek}.
\end{itemize}

\paratitle{Post-training with Multi-Stage SFT, Distillation, and Reinforcement Learning.}

\begin{itemize}[leftmargin=10pt]
    \item \textbf{Multi-Stage SFT.} We first fine-tune our base model on over 30 million cleaned samples related to math, science, and code to establish strong reasoning capabilities. Subsequently, we apply curriculum learning with increasingly diverse and more challenging instructions to ensure robust performance across a wide range of tasks. To further improve the quality of SFT responses, we introduce an innovative approach that combines deliberative learning with Chain-of-Thought (CoT) reconstruction: we first refine the Solution component to make it superior, then reconstruct a corresponding CoT that logically leads to this improved solution. This yields high-quality SFT training examples that significantly outperform those generated via rejection sampling.

    \item \textbf{Distillation.} Following SFT, we employ the Nanbeige flagship reasoning model as the teacher to distill the Nanbeige4-3B student model, with our proposed Dual Preference Distillaition (DPD) method. In particular, we innovate the loss function design: on the one hand, the student model learns to mimic the teacher’s output distribution as a policy model; on the other hand, it is simultaneously trained to distinguish between high-quality and low-quality responses.

    \item \textbf{Reinforcement Learning.} Building upon the distilled model, we conduct multi-stage RL training to further boost performance. We design a suite of reward and verification strategies tailored to different training phases—including STEM reasoning, coding, and human preference alignment—ensuring consistent and stable improvements throughout.
\end{itemize}

Building on the techniques described above, we develop a base model Nanbeige4-3B-Base, and a reasoning-enhanced model Nanbeige4-3B-Thinking. To support the community and foster open, reproducible research in LLMs, we \textbf{open-source the Nanbeige4-3B model family}. We hope this will empower researchers and developers to explore advanced training methodologies, accelerate innovation in reasoning-centric models, and contribute to a more open and collaborative AI ecosystem.
\section{Pre-Training}
\label{sec:pretraining}
In this section, we first describe the construction of our pre-training dataset and the corresponding training recipe, followed by an evaluation of the Nanbeige4-3B-Base model’s performance.

\subsection{Pre-Training Data}

During the construction of the pre-training corpus for Nanbeige4-3B, we extensively collect high-quality and diverse data.
Our corpus encompasses a diverse range of web pages, scholarly articles, books, source code, and other materials.  
To strengthen the grasp of the human world, we not only extract clean text from HTML documents but also develop a highly efficient PDF text-extraction pipeline.  
We further augment the pre-training mix with synthetic data that target specialized competencies: these take the form of question–answer pairs, textbooks, lecture notes, and long chain-of-thought samples, which constitute 15\% of the total pre-training tokens.

\subsection{Pre-Training Recipe}
In developing the training recipe for Nanbeige4-3B, we focus on two key issues:
\begin{enumerate}[label=\textbf{\arabic*.}, leftmargin=*, nosep]
  \item How to identify high-quality data while filtering out low-quality samples.
  \item How to make full use of the selected data to further improve performance.
\end{enumerate}
In the subsequent parts of this section, we first introduce how data quality is assessed and annotated, and then present how these quality ratings are utilized during training to enhance model performance.

\subsubsection{Data Quality Identify}
To retain high-quality samples while filtering out low-quality ones, we access the quality of each data entry in the corpus using two complementary strategies: (1) multi-dimensional tagging that inspects intrinsic attributes, and (2) similarity-based scoring against a curated set of high-quality seed data to assess extrinsic alignment.

\paratitle{Multi-dimensional tagging.}
We adopt a similar workflow as in prior research on pre-training data quality assessment \cite{hu2024minicpmunveilingpotentialsmall, su2025nemotroncctransformingcommoncrawl}. This involves constructing a comprehensive labeling system, sampling and annotating data using a strong model, distilling the annotations to a smaller model for scalable annotation, and finally applying weighted ranking to select high-quality samples. Our labeling framework spans two key aspects: format and content. 
Initially, we define over 60 dimensions, including knowledge density, reasoning density, and text fluency, etc. 
Based on fine-grained experiments and multiple criteria such as inter-dimension similarity, we further filter these candidates and ultimately retain a final set of 20 dimensions.
We annotate all pre-training data along these 20 dimensions and further perform extensive validation and optimization experiments to refine their effectiveness.
Empirical analyses reveal two key findings. First, content-related labels are markedly more predictive of data quality than format-related ones. Second, a fine-grained 0–9 scoring scheme provides substantially more accurate data selection than a binary 0–1 labeling strategy.

\paratitle{Similarity-based scoring.}
We build a retrieval database containing hundreds of billions of entries, supporting efficient hybrid text-based and vector-based retrieval. With this robust retrieval infrastructure, we continuously iterate on seed data curation, as well as the retrieval methodologies.
For the seed data, we prioritize samples that rank high within our quality tagging framework while ensuring data provenance from reliable and authoritative sources. This dual-criteria approach guarantees that our foundational dataset maintains both exceptional quality standards and trustworthy origins. For the retrieval methodology, we conduct extensive experiments on how to balance between similarity scores and quality assessments. We discover that applying retrieval strategies on top of our quality labeling system enables more precise identification and selection of high-quality data.

By combining multi-dimensional tagging with similarity-based scoring, we filter out tens of trillions of tokens that do not meet our criteria and
retain 12.5 trillion high-quality training data. From these 12.5 tokens, we further select 6.5 trillion tokens of even higher quality for up-sampling two or more epochs, ultimately forming our final 23 trillion token training corpus.

\subsubsection{Data Utility Scheduler}

To fully leverage the selected data for further performance improvement, we introduce a novel learning rate scheduler named Fine-Grained Warmup-Stable-Decay~(FG-WSD).
This scheduler increases the learning rate during an initial warm-up phase, maintains it across multiple carefully designed stable stages, and finally applies a smooth decay.
The multi-stage stable phases provide greater room for exploration based on the previously obtained data scores, while preventing the undesirable coupling between data ordering and learning rate changes.

\paratitle{Cosine Decay vs. WSD vs. FG-WSD}
Pre-training learning rate schedulers generally fall into two categories: warmup-cosine-decay~\cite{touvron2023llamaopenefficientfoundation, touvron2023llama2openfoundation, grattafiori2024llama3herdmodels} and warmup-stable-decay~(WSD)~\cite{hu2024minicpmunveilingpotentialsmall, liu2024deepseek, minimax2025minimaxm1scalingtesttimecompute}. Through experimental validation, we find that when the data quality during the annealing phase is sufficiently high, warmup-stable-decay significantly outperforms warmup-cosine-decay. Consequently, we adopt the WSD approach as our foundational scheduling strategy.
Building upon this baseline, we further introduce an enhanced variant: FG-WSD (Fine-Grained WSD). Recognizing that the WSD schedule maintains a constant learning rate during the stable phase, we optimize the data resampling strategy accordingly. Rather than uniformly sampling high-quality data throughout the entire training process, FG-WSD divides training into multiple fine-grained stages and progressively increases the proportion of higher-quality data mixtures in later stages.

\paratitle{Preliminary Experiments.}
We verify the effectiveness of FG-WSD on a 1B-parameter model with a fixed 100B-token corpus in the decay stage. The stable phase consumes one epoch of 500B medium-quality (MQ) tokens and two epochs of 250B high-quality (HQ) tokens, for a total of 1T stable-phase tokens.  Under vanilla WSD, these 1T tokens are shuffled uniformly: every sample is drawn from a static 1:1 mixture of HQ and MQ data. FG-WSD, by contrast, splits the stable phase into two contiguous stages.  Stage 1 processes 750B tokens composed of one epoch of HQ data (250B tokens) plus one epoch of MQ data (500B tokens).  Stage 2 then continues for an additional 250B-token containing only HQ subset.

\begin{table}[!t]
\centering
\caption{Performance comparison between vanilla WSD scheduler and our proposed Fine-Grained WSD scheduler. The experiment is implemented on 1B parameter scale model with 1T tokens.}
\label{wsd}
\resizebox{0.92 \textwidth}{!}{
\begin{tabular}{lcccccc}
\toprule
\centering
Learing Rate Scheduler &GSM8k & Cmath & BBH &MMLU & CMMLU &  MMLU-Pro \\
\midrule
Vanallia WSD & 27.1 & 34.5 & 29.3 & 49.2& 50.3 & 16.87\\
Fine-Grained WSD & \textbf{34.3} & \textbf{39.5} & \textbf{31.6}  & \textbf{50.6} & \textbf{51.9} & \textbf{18.64} \\
\bottomrule
\end{tabular}
}
\end{table}

\begin{table}[!t]
\centering
\caption{Training stages of the FG-WSD scheduler in Nanbeige4-3B-Base, with corresponding data usage and learning rates.}
\label{tab:training_stages}
\resizebox{0.72 \textwidth}{!}{
\begin{tabular}{lcc}
\toprule
{Stage} & {Training Tokens} & {Learning Rate} \\
\midrule
Warmup Stage & 0.1T & 0 $\rightarrow$ 4.5e-4 \\
Diversity-Enriched Stable Stage & 12.4T & Constant 4.5e-4 \\
High-Quality Stable Stage & 6.5T & Constant 4.5e-4 \\
Decay Stage & 4T & 4.5e-4 $\rightarrow$ 1.5e-6 \\
\bottomrule
\end{tabular}
}
\end{table}

As shown in Table~\ref{wsd}, our FG-WSD method outperforms the vanilla WSD approach across all benchmarks\footnote{For GSM8k, Cmath, and BBH, we run 3-shot evaluation. For MMLU, CMMLU, and MMLU-Pro, we run 5-shot evaluation.} Notably, the improvement is more pronounced on mathematical and reasoning tasks (e.g., GSM8K~\cite{cobbe2021trainingverifierssolvemath}, CMATH~\cite{wei2023cmathlanguagemodelpass}, and BBH~\cite{suzgun2022challengingbigbenchtaskschainofthought}) compared to knowledge and science benchmarks (e.g., MMLU~\cite{hendrycks2021measuringmassivemultitasklanguage}, CMMLU~\cite{li2024cmmlumeasuringmassivemultitask}, and MMLU-Pro~\cite{wang2024mmluprorobustchallengingmultitask}). This is because our quality scoring framework prioritizes reasoning density over knowledge density during high-quality data selection.


After validating the effectiveness of FG-WSD using a 1B-parameter model trained on 1T tokens, we scale the method up to the full training corpus of Nanbeige4-3B-Base. The overall training pipeline is organized into four well-defined stages: Warmup, Diversity-Enriched Stable, High-Quality Stable, and Decay. The amount of training tokens allocated to each stage, as well as the corresponding learning-rate schedules, are summarized in Table~\ref{tab:training_stages}.
A noteworthy aspect of the final decay stage is the use of the ABF (Adjusting Base Frequency) method~\cite{xiong2023effectivelongcontextscalingfoundation}, which enables us to extend the model’s context length to 64K. This long-context capability allows the model to fully ingest synthetic long chain-of-thought traces, books, academic articles, and large-scale code repositories without truncation, ensuring that high-value long-form data is preserved throughout training.

\subsection{Post-SFT Evaluation for Base Model}
Beyond evaluating base-model capabilities through few-shot evaluation, post-SFT downstream performance offers a more informative indicator, as virtually all production systems ultimately rely on instruction-tuned variants.
We therefore fine-tune Nanbeige4-3B-Base and the open-source Qwen3-Base series with an identical supervised fine-tuning (SFT) pipeline and compared the resulting checkpoints on six representative benchmarks: AIME~2024, AIME~2025, GPQA~\cite{rein2023gpqagraduatelevelgoogleproofqa}, MATH-500~\cite{lightman2023letsverifystepstep}, LiveCodeBench-V5, and LiveCodeBench-V6~\cite{jain2024livecodebenchholisticcontaminationfree}.

To ensure robustness, we repeat the procedure with three independent SFT datasets, including Nemotron-Post-Training-Dataset-V1~\cite{NemotronPostTrainingDatasetV1}, Ring-Lite-SFT-Data~\cite{ringteam2025ringlitescalablereasoningc3postabilized}, and OpenThoughts-3~\cite{guha2025openthoughtsdatarecipesreasoning}, aggregating 9 SFT training runs in total. For each dataset, we randomly sample 500,000 instances and train for 2 epochs. To ensure a rigorously controlled and fair comparison, we keep all training hyperparameters identical across the different base models.

\begin{table}[!t]
\centering
\caption{Comparison of base models on reasoning benchmarks after fine-tuning.}
\label{tab:combined}
\resizebox{1.0\textwidth}{!}{
\begin{tabular}{llcccccc}
\toprule
SFT Dataset & Base Model & AIME~24 & AIME~25 & Math-500 & GPQA & LCB-V5 & LCB-V6 \\
\midrule

\multirow{3}{*}{Nemotron-Post-Training-V1} 
& Qwen3-4B-Base & 24.6 & 25.0 & 90.4 & 44.6 & 15.9 & 17.0 \\
& Qwen3-8B-Base & \underline{37.9} & \underline{29.6} & \underline{91.1} & \underline{48.9} & \underline{27.6} & \underline{28.1} \\
& Nanbeige4-3B-Base & \textbf{52.9} & \textbf{40.8} & \textbf{93.4} & \textbf{53.4} & \textbf{35.9} & \textbf{34.0} \\
\midrule

\multirow{3}{*}{Ring-Lite-SFT} 
& Qwen3-4B-Base & 40.4 & 31.3 & 93.6 & 51.4 & 20.7 & 22.5 \\
& Qwen3-8B-Base & \underline{50.0} & \underline{35.8} & \underline{94.4} & \underline{55.1} & \underline{30.2} & \underline{29.5} \\
& Nanbeige4-3B-Base & \textbf{56.8} & \textbf{45.3} & \textbf{95.5} & \textbf{57.7} & \textbf{33.3} & \textbf{33.2} \\
\midrule

\multirow{3}{*}{Openthoughts3} 
& Qwen3-4B-Base & 52.9 & 42.1 & 93.2 & 49.6 & 27.2 & 27.5 \\
& Qwen3-8B-Base & \underline{60.4} & \underline{47.1} & \textbf{95.0} & \underline{55.3} & \underline{35.2} & \underline{34.4} \\
& Nanbeige4-3B-Base & \textbf{62.4} & \textbf{49.2} & \underline{94.6} & \textbf{56.9} & \textbf{40.9} & \textbf{38.8} \\
\bottomrule
\end{tabular}
}
\end{table}


As shown in Table~\ref{tab:combined}, across all experimental settings, Nanbeige4-3B-Base not only delivers substantial improvements over Qwen3-4B-Base, but also consistently outperforms the much larger Qwen3-8B-Base, despite being only half its size. These results indicate that our model provides a markedly stronger initialization for developing downstream reasoning models.
\section{Post-Training}

\begin{figure}[!h] 
    \centering
    \includegraphics[width=1\textwidth]{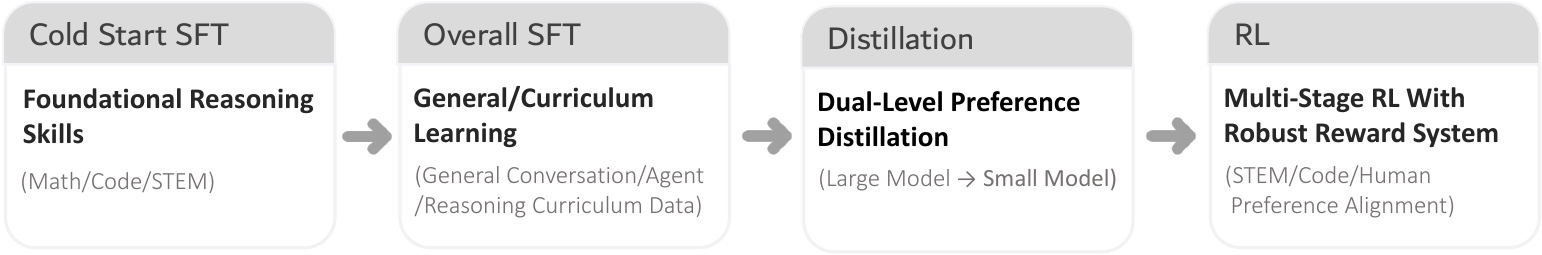} 
    \caption{Nanbeige4-3B-Thinking Post-Training Pipeline} 
    \label{fig:nbg_pipeline} 
\end{figure}

In this section, we introduce each stage of the post-training process in detail.
The overall post-training pipeline for Nanbeige4-3B-Thinking is illustrated in Figure~\ref{fig:nbg_pipeline}. Initially, we conduct two fine-tuning stages—cold-start SFT and overall SFT—to equip the model with fundamental reasoning capabilities across a wide range of tasks. Subsequently, a knowledge-distillation phase is employed to further enhance performance. Finally, the model undergoes multi-stage RL training to achieve additional improvements in multiple domain skills.

\label{sec:posttraining}

\subsection{Cold Start Supervised Fine-tuning}

On top of the base model, we first perform a cold-start supervised fine-tuning (SFT) stage to establish a robust foundation for reasoning. This stage focuses on high-quality reasoning data. After systematic cleaning and filtering, we collect approximately 30 million QA samples of mathematical, code, and subject-area problem-solving and reasoning. Based on this dataset, we construct a training corpus with a context length of 32K tokens, comprising approximately 50\% mathematical reasoning, 30\% scientific reasoning, and 20\% code-related tasks. The objective of the cold-start stage is to strengthen the model's chain-of-thought reasoning and structured response abilities~\cite{wei2022chainofthought,wang2022selfconsistency}, providing a solid foundation for subsequent capability expansion.

\begin{figure}[!h]
  \centering
  \includegraphics[width=0.8\textwidth]{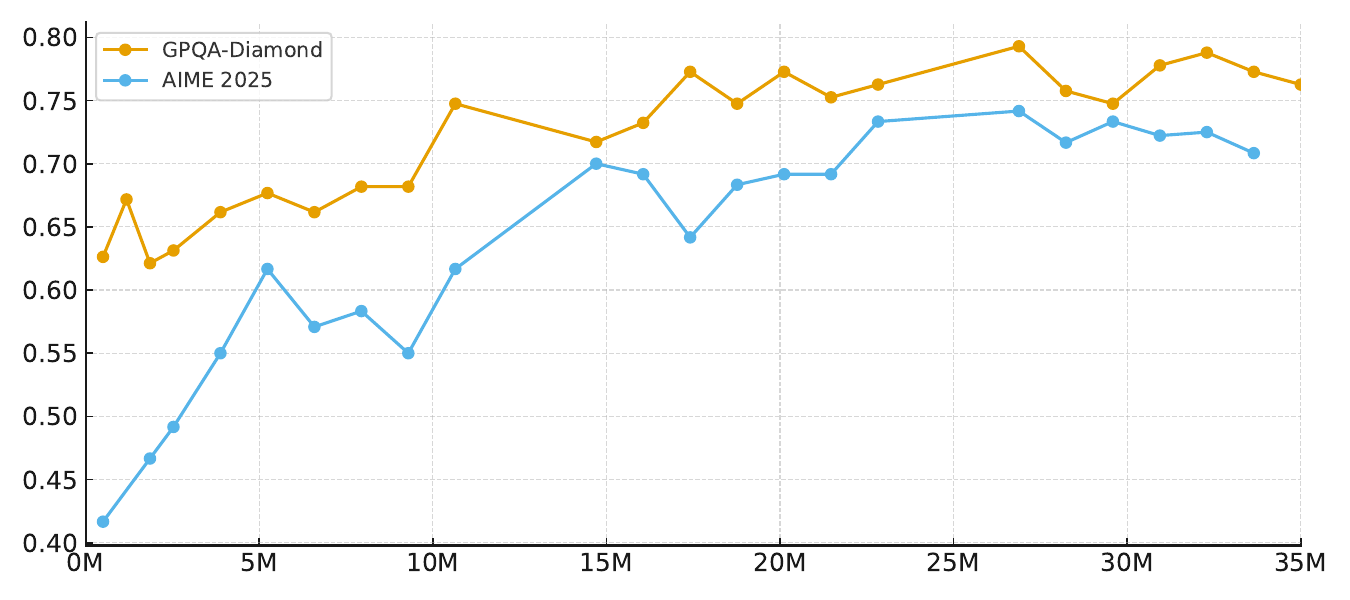}
  \caption{
    Cold-start SFT scaling (0.5M–35M) on NBG4-3B reasoning benchmarks.
  }
  \label{fig:coldstart-scaling}
\end{figure}

\paratitle{Scaling SFT Instructions.}
While some recent technical reports suggest that tens or hundreds of thousands of high-quality instructions are sufficient for supervised-finetuning, our experiments on Nanbeige4-3B reveal a different trend. 
As shown in Figure~\ref{fig:coldstart-scaling}, when controlling for data distribution and overall quality, scaling the cold-start SFT instruction set from hundreds of thousands to tens of millions of examples continues to produce substantial improvements on challenging reasoning benchmarks such as AIME~2025 and GPQA-Diamond, without a clear early saturation point. 
Motivated by this empirical observation, we explicitly adopt a Scaling SFT Instructions strategy. Rather than relying on a compact instruction set, we train the model during the cold-start stage on tens of millions of carefully curated instructions. This large-scale instruction regime provides the model with a stronger initial reasoning prior and stabilizes its chain-of-thought generation policy, thereby creating a more reliable foundation for subsequent training stages.

\subsection{Overall Supervised Fine-Tuning}

After the model acquires initial reasoning capabilities, we perform an Overall SFT stage~\cite{ouyang2022instructgpt,wang2022selfinstruct,zhou2023lima} to further enhance its general abilities and task diversity. The training corpus combines general conversation and writing data (covering everyday dialogue and multiple genres), agent-style interaction data (tool use, task decomposition, planning, and execution), harder reasoning data that targets the weak spots revealed in the cold-start stage, and code-related tasks that reinforce programming and code understanding. Using a 64K context length, we mix the data as roughly 40\% mathematical and subject-specific reasoning, 30\% general QA and writing, 20\% agent scenarios, and 10\% coding tasks, which systematically improves the model’s general dialogue ability, task execution, and adaptation to diverse application scenarios while maintaining strong reasoning performance.
In the following section, we provide a detailed description of our data processing pipeline.

\begin{figure}[!h] 
    \centering
    \includegraphics[width=1\textwidth]{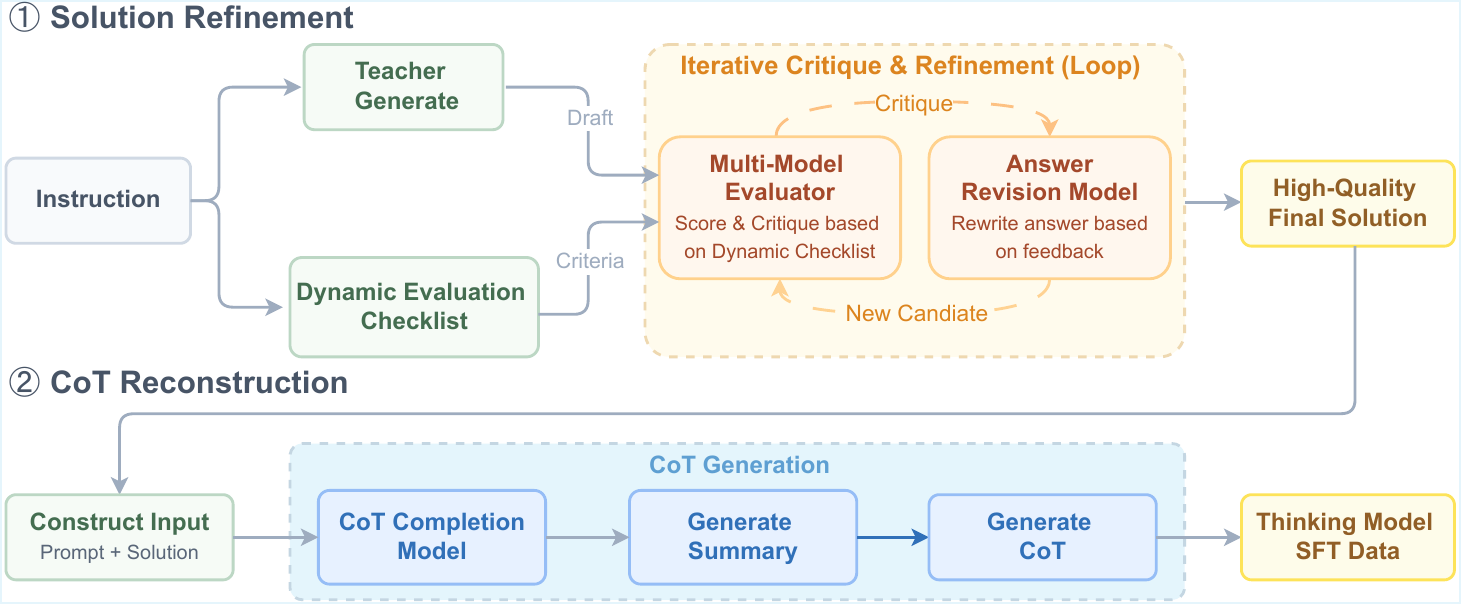} 
    \caption{Deliberative generation refinement and CoT completion} 
    \label{fig:Deliberative} 
\end{figure}

\paratitle{Solution Refinement.}
To enhance the model’s overall output quality on complex tasks, we develop a unified mechanism that combines deliberative generation refinement with chain completion~\cite{tencent2025hunyuanturbos}. Specifically, for each instruction, the system first constructs a tailored multi-dimensional evaluation checklist by selecting appropriate criteria from a predefined pool according to the instruction’s semantics and task type, and then adopts this checklist as an explicit constraint for the current evaluation round. The checklist selectively integrates criteria such as correctness, completeness, consistency, executability, and safety, and further refines each criterion into concrete checkpoints to enable fine-grained assessment of candidate responses.
In the next iteration, the system dynamically selects from a pool of teacher models the one that performs best on the current instruction and has it co-generate candidate answers with the current SFT/Thinking model. An evaluation model then conducts cross-evaluation and comparative scoring of all candidates against a predefined checklist, producing structured feedback that highlights error locations, missing steps, and optimization suggestions. This feedback drives the Thinking model through iterative generate–review–revise cycles, continuously improving the quality of the solution.

\paratitle{CoT Reconstruction.}
After multiple rounds of deliberation and rewriting, although the final solution quality is greatly improved, the original chain of thought is often disrupted or lost, making it difficult to obtain supervision signals that simultaneously provide a high-quality final answer and a stable, learnable reasoning process.  To address this, we additionally train a chain-completion model.  When constructing training data for the Thinking model, we feed this model with the concatenation of the instruction and the final solution after multi-round refinement.  Empirically, generating a summarized first improves followability, so the model first generates a brief summary chain of thought and then produces an explicit chain of thought that is consistent with the final answer.  Finally, we concatenate the completed chain of thought and the final answer as the target output and use it as the training sample for the Thinking model, thus restoring a structured and well-aligned reasoning supervision signal while preserving answer quality.

After integrating the above mechanism into the Overall SFT stage, the model's alignment with human preferences is significantly improved: on the Arena-Hard v2 benchmark, the score improves by 16\%, with no degradation in reasoning capability~\cite{li2024arenahard}.

\paratitle{Function Call Supporting.}
Our model provides native support for the function-call paradigm, enabling seamless tool invocation through formally defined and standardized parameter specifications. During the overall SFT stage, we proportionally increase the amount of function-call (FC) data to further strengthen this capability.
To enhance the model’s proficiency in function calling, we incorporate both open-source and synthetic data. For the open-source portion, we unify data formats and reconstruct responses so that each sample strictly adheres to our function-call schema and includes an explicit reasoning path. For the synthetic portion, we deploy real-world environments and leverage strong models (e.g., Nanbeige3.5-Pro) to generate high-quality trajectories via rejection sampling. In addition, we adopt a multi-agent framework to simulate realistic user–assistant–tool interactions, creating data that span diverse scenarios and a wide range of difficulty levels~\cite{yang2025toolmind}.

\begin{figure}[t] 
    \centering
    \includegraphics[width=1\textwidth]{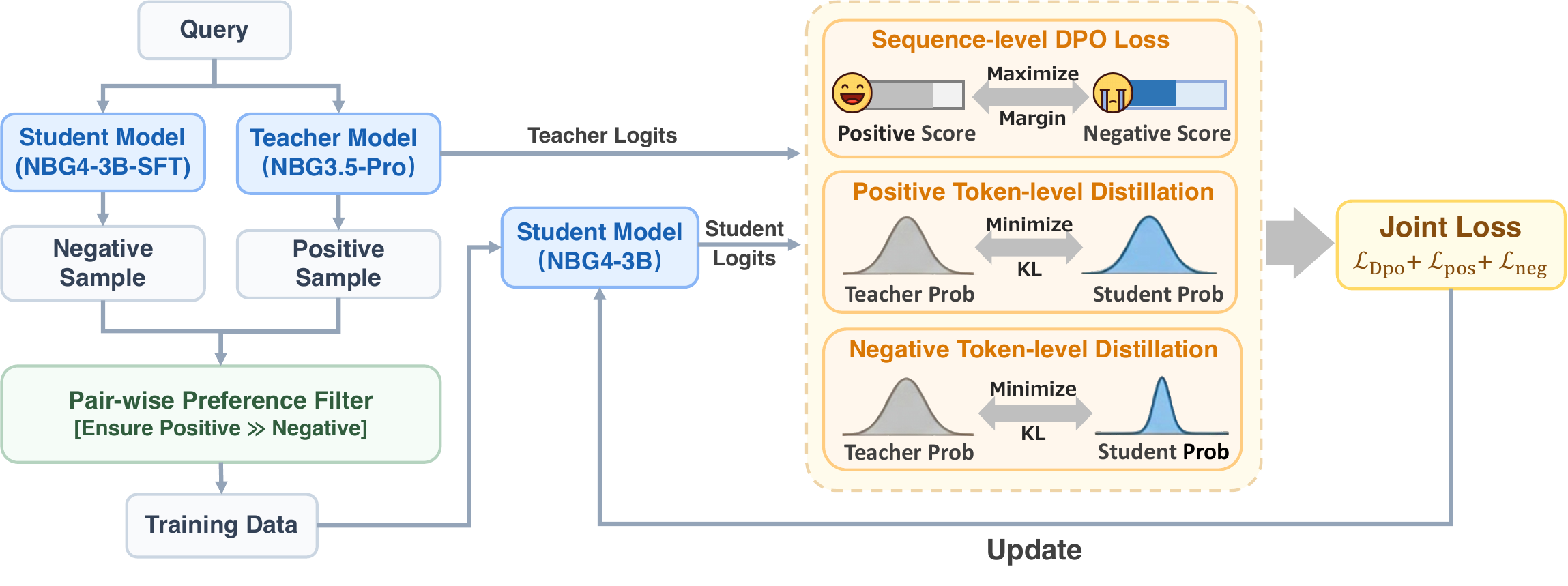} 
    \caption{Overview of the Dual-Level Preference Distillation (DPD) framework} 
    \label{fig:Distillation} 
\end{figure}

\subsection{Dual-level Preference Distillation}

We propose \textbf{Dual-Level Preference Distillation (DPD)}, a joint training framework that harmonizes token-level knowledge distillation with sequence-level preference optimization. In this framework, Direct Preference Optimization (DPO)~\cite{rafailov2023dpo} acts as a sequence-level decision-boundary regularizer, maximizing the margin between positive and negative responses. Simultaneously, we introduce token-level supervision from the teacher model's probability distribution on both sample types. This dual-granularity setup improves instruction-following behavior and subjective preference alignment, and at the same time, substantially enhances the model's complex reasoning capabilities.

For data construction, positive samples are obtained by sampling multiple responses from the teacher model Nanbeige3.5-Pro for the same instruction, followed by model-based scoring and rule-based filtering to select the highest-scoring answer. Negative samples are generated by sampling from the 3B student model under training. These candidates are then passed through automatic evaluation and rule-based checks, and only those whose quality and scores are significantly worse than the corresponding positive sample are retained.

For the optimization objective, we use a joint loss that combines token-level probability distillation with a sequence-level DPO preference loss.  On positive samples, the student model is trained to match the Nanbeige3.5 Pro probability distribution at each token.  On negative samples, we also apply a distillation loss, where the teacher provides a reference distribution for the incorrect responses generated by the student, reducing the probability of highly confident erroneous tokens and increasing the probability of under-estimated but reasonable alternatives.  This design enhances the model’s ability to correct its own mistakes and to recognize errors.  The sequence level DPO preference loss, implemented as a margin constraint, explicitly enlarges the score gap between positive and negative responses, thereby sharpening the decision boundary and improving style alignment.

Using the Nanbeige4-3B SFT model as the baseline, training with this framework yields consistent and substantial relative gains on multiple internal and public benchmarks: around an 8\% improvement on challenging mathematical benchmarks such as AIME~2024 and AIME~2025~\cite{emergentmind2025aime}, about a 10\% gain on scientific reasoning tasks such as GPQA~\cite{rein2023gpqa}, and roughly a 30\% improvement on BFCL V4, which emphasize tool use and problem decomposition~\cite{patil2025bfcl}. At the same time, Arena-style subjective preference evaluations still show an improvement of around 8\%~\cite{li2024arenahard}. 
Furthermore, incorporating an RL phase on top of this distillation framework yields substantially larger gains compared to initiating RL directly from the SFT baseline.

\subsection{Reinforcement Learning}
Large-scale supervised fine-tuning and distillation provide a strong foundation for the reasoning ability of NBG4-3B. We further bootstrap both its reasoning capability and human preference alignment through reinforcement learning (RL). Instead of conducting a single-stage RL procedure on a mixed corpus spanning multiple knowledge domains~\cite{yang2025qwen3}, we adopt a multi-stage RL framework, where each stage targets a specific ability dimension. To fully exploit the potential of each stage, we perform on-policy data filtering with the latest model before every RL phase. For each phase, we adopt appropriate reward models or verifiers based on the features of the training data.

\subsubsection{On-Policy Data Filtering}
During multi-stage RL training, the model's reasoning ability improves after each stage. Due to cross-domain knowledge transfer, samples that were previously challenging may become trivial in later stages and contribute little to further improvement. To maintain a high-quality learning signal, we apply on-policy data filtering before each RL stage using the model from the previous stage.

Concretely, we use the model from the preceding stage to compute the avg@16 accuracy for every question and retain only those samples whose pass rate lies strictly between 10\% and 90\%. This focuses training on problems that are neither trivial nor unsolvable, thereby maximizing the effectiveness of each RL update.

\subsubsection{Multi-Stage Reinforcement Learning}
To enhance reasoning ability across domains, we employ multi-stage RL rather than a single mixed-corpus paradigm. While mixed-corpus training can yield strong cross-domain transfer~\cite{cheng2025revisitingreinforcementlearningllm}, we observe that it often slows progress in specific domains: the model may spend many updates to achieve marginal gains in more challenging skills. For example, when jointly training on advanced mathematics and competitive programming data, the model tends to improve more on mathematics than on competitive coding. To address this imbalance, we organize RL into multiple stages, each focused on a single domain. This allows the model to concentrate its capacity on domain-specific skills at each stage while still benefiting from cross-domain transfer over the full training pipeline.

We adopt on-policy GRPO~\cite{shao2024deepseekmathpushinglimitsmathematical} for each RL stage, with several enhancements for stablized training. In particular, we remove the KL penalty term and mask the loss for truncated sequences, following the insights of DAPO~\cite{yu2025dapoopensourcellmreinforcement}. We organize the training into three RL stages as follows.

\paratitle{STEM RL with Agentic Verifier.}
We use STEM-focused RL as the first stage, motivated by prior findings that STEM training provides strong cross-domain transfer. Our dataset consists of question–answer pairs in mathematics and the natural sciences. For mathematical problems, we collect diverse data from open-source datasets. For science problems spanning physics, chemistry, and biology, we use proprietary competition-level collections. For problems with multiple sub-parts, we rewrite each sub-problem into a self-contained question with a complete context.

In STEM domains, reference answers and model output may express equivalent numerical results in different symbolic forms. To provide accurate training signals, we employ a tool-augmented verifier that calls Python interpreter to perform exact computation and simplification~\cite{CosineVerifier}. This agentic verifier enables robust judgments that go beyond string-matching rules.

\paratitle{Practical Coding RL with Synthetic Test Functions.}
This stage aims to enhance practical coding ability across multiple programming languages and task scenarios. We design a multi-agent system to synthesize problems paired with executable test functions. To ensure the correctness and completeness of the synthetic data~\cite{chou2025autocodebenchlargelanguagemodels}, we adopt a reverse-generation procedure: we first synthesize the solution and its corresponding test functions, and only then generate the natural-language problem description. Concretely, it first retrieves high-quality code snippets from GitHub, then refines or evolves these snippets into self-contained, verifiable solutions and produces paired public and private test functions. Finally, all candidate triples (problem, solution, test function) are validated via sandboxed execution to guarantee reliability. During RL training, these test functions are executed to provide a binary reward signal based on whether the generated solution passes all tests.

\paratitle{Human Preference Alignment RL with Pairwise Reward Model.}
In the final stage, we focus on aligning the model with human preferences on tasks such as creative writing and role-playing. Since these tasks prefer open-ended responses without fixed reference answers, many prior works rely on general-purpose language models to score the human preference alignment of candidate responses~\cite{li2024crowdsourced}. However, using a general language model as a reward model has two key drawbacks: (i) it often requires a lengthy chain of thought before reaching a final verdict, which is highly time-consuming; and (ii) RL training is prone to reward hacking. To address these issues, we train a pairwise reward model that can express preferences using only a few tokens, while exhibiting strong resistance to reward hacking.

During reinforcement learning (RL) training, we first sample diverse instructions that are both challenging and clearly specified. We then prompt strong baseline models (e.g., Nanbeige3.5-Pro) to generate high-quality reference responses. Each rollout produced by the policy model Nanbeige4-3B is paired with its corresponding reference response, and the pairwise reward model assigns a preference-based score that serves as the reward signal for policy optimization.

\subsection{Post-Training Evaluation}

\begin{table}[t]
\centering
\small
\setlength{\tabcolsep}{3pt} 
\caption{Comparison between Nanbeige4-3B-Thinking and Qwen series reasoning models.}
\label{tab:nbg4-3b-evaluation}
\begin{tabular}{lcccccc}
\toprule
Benchmark
& \makecell[c]{Qwen3-4B-\\2507}
& \makecell[c]{Qwen3-8B-\\2504}
& \makecell[c]{Qwen3-14B-\\2504}
& \makecell[c]{Qwen3-30A3-\\2507}
& \makecell[c]{Qwen3-32B\\2504}
& \makecell[c]{ Nanbeige4-3B-\\2511} \\
\midrule
\multicolumn{7}{c}{Mathematical Reasoning} \\
\midrule
AIME2025      &81.3 & 67.3 & 70.4 &\underline{85.0} &72.9&\textbf{85.6}   \\
AIME2024          & 83.3 & 76.0 & 79.3 &\underline{89.2} & 81.4&\textbf{90.4}  \\
\midrule
\multicolumn{7}{c}{Scientific Reasoning} \\
\midrule
GPQA-Diamond      & 67.2 & 62.0 & 64.0 &\underline{73.4} &68.7&\textbf{82.2}   \\
SuperGPQA          & 46.7 & 39.1 & 46.8 &\textbf{56.8} &\underline{54.1} & 53.2   \\
\midrule
\multicolumn{7}{c}{{Tool Use \& Coding}} \\
\midrule
BFCL-V4      & 44.9 & 42.2 & 45.4 & \underline{48.6} & 47.9&\textbf{53.8}  \\
Fullstack Bench         & 47.1 & 51.5 & 55.7 &\underline{54.4}& \textbf{58.2}& 48.0  \\ 
\midrule
\multicolumn{7}{c}{{Human Preference Alignment}} \\
\midrule
ArenaHard-V2     & 40.5 & 26.4 & 39.9 & \textbf{60.0}&48.4& \textbf{60.0}  \\
Multi-Challenge     & \underline{41.8} & 35.8 & 36.4 &\textbf{49.4}&39.2& 41.2  \\
\bottomrule
\end{tabular}
\end{table}
We evaluate Nanbeige4-3B on a diverse suite of benchmarks to more comprehensively assess its performance. For each benchmark, we conduct multiple evaluation runs and report the average score across these repeated trials (e.g., avg@8 denotes the mean score computed over eight independent repetitions). Our evaluation encompasses the following benchmarks:

\begin{itemize}[leftmargin=10pt]
    \item \textbf{Mathematical Reasoning:} AIME~2025, AIME~2024~\cite{emergentmind2025aime} (reported as avg@8)
    \item \textbf{Scientific Reasoning:} GPQA-Diamond~\cite{rein2023gpqagraduatelevelgoogleproofqa}, superGPQA~\cite{pteam2025supergpqascalingllmevaluation} (avg@3)
    \item \textbf{Tool Use \& Coding:} BFCL-V4~\cite{patil2025bfcl}, Fullstack-Bench~\cite{bytedanceseedfoundationcodeteam2025fullstackbenchevaluatingllms} (avg@3)
    \item \textbf{Human Preference Alignment:} Arena-Hard V2~\cite{li2024crowdsourced}, Multi-Challenge~\cite{sirdeshmukh2025multichallengerealisticmultiturnconversation} (avg@3)
\end{itemize}

For all benchmarks, we use a sampling temperature of 0.6 and top-p of 0.95, and we set the maximum generation length to 64k tokens. We compare Nanbeige4-3B against a series of open-source small language models: Qwen3-4B, Qwen3-8B, Qwen3-14B, Qwen3-30B-A3B, and Qwen3-32B. We report metrics of other models with recommended hyperparameters in our environment. Table~\ref{tab:nbg4-3b-evaluation} presents the evaluation results.

Experimental results show that Nanbeige4-3B excels at demanding reasoning tasks despite its compact size. It sets new state-of-the-art averages on AIME 2024, AIME 2025 and GPQA-Diamond, outperforming models with up to 10× more parameters, including Qwen3-32B and Qwen3-30B-A3B.
Beyond mathematics and science, the model demonstrates strong tool-use proficiency, scoring 53.8 on BFCL-V4—an absolute gain of 5.2 points over Qwen3-30B-A3B. Human-preference alignment is equally notable: Nanbeige4-3B matches the 60.0 point top score on Arena-Hard V2, yielding a 50\% relative improvement over Qwen3-4B.

In addition to the benchmark results from our locally deployed evaluations, Nanbeige4-3B also demonstrates outstanding performance in external public evaluations. As shown in Table~\ref{tab:writingbench}, Nanbeige4-3B-Thinking-2511 ranks among the top models on the WritingBench Leaderboard (November 2025), showcasing writing capabilities across diverse scenarios that are on par with much larger models.

\section{Conclusion}
In this work, we introduce Nanbeige4-3B, a compact yet highly capable 3-billion-parameter language model that redefines what small-scale models can achieve through innovation on data and training paradigm. Trained on a meticulously curated 23 trillion tokens of high-quality data and enhanced with novel post-training techniques—including chain-of-thought introspection and refinement, advanced reinforcement learning, and knowledge distillation—Nanbeige4-3B demonstrates remarkable reasoning and generation capabilities across diverse domains.

Despite its modest size, Nanbeige4-3B outperforms open-source counterparts such as Qwen3-8B and Qwen3-14B on challenging benchmarks, including AIME, SuperGPQA, Arena-Hard V2, and BFCL-V4. Notably, in the latest official WritingBench leaderboard, the Nanbeige4-3B-Thinking-2511 achieves top-tier performance, rivaling that of large-scale models like Deepseek-R1-0528. Nanbeige4-3B thus establishes a new lightweight flagship paradigm, offering high capability, efficiency, and accessibility for both research and real-world deployment.

Looking ahead, we aim to further extend the capabilities of small-scale models to even more complex challenges, such as autonomous software engineering (SWE), deep-research agents, and diverse real-world cross-scenario tool use tasks.

\newpage

\bibliographystyle{plain} 
\bibliography{references}
\newpage
\appendix
\section{Author List}

Authors are listed in \textbf{alphabetical order by first name}. Names marked with an asterisk (*) denote individuals who were previously affiliated with our team. Yang Song is the corresponding author and can be reached at \texttt{songyang@kanzhun.com}.

Chen Yang, Guangyue Peng, Jiaying Zhu, Ran Le, Ruixiang Feng, Tao Zhang, Wei Ruan,  Xiaoqi Liu\textsuperscript{*}, Xiaoxue Cheng, Xiyun Xu, Yang Song\textsuperscript{\dag}, Yanzipeng Gao\textsuperscript{*}, Yiming Jia, Yun Xing, Yuntao Wen, Zekai Wang, Zhenwei An, Zhicong Sun, Zongchao Chen

\end{document}